\documentclass[letterpaper, 10 pt, journal, twoside]{IEEEtran}

\usepackage{times}

\usepackage[bookmarks=true]{hyperref}
\usepackage[numbers]{natbib}
\usepackage{multicol}

\usepackage{graphicx}
\usepackage{subcaption}
\usepackage{xcolor}
\usepackage{amssymb}
\usepackage{amsmath}
\usepackage{amsthm}
\usepackage{booktabs}
\usepackage[percent]{overpic}
\usepackage[bbsets,Dfprime]{math}
\usepackage[prefixflatinterpret,bracketmodalinterpret,modernsign,substopindex,shortmquant,mquantifiertype,mconnectiveformal,bracketinterpret,fixformat,setfixinterpret,modifopindex,seqarrow,seqoptional,sidenotecalculus,abbrseqcontext,shortterms,nosigmaterms,novarterms]{logic}
\usepackage[pretest,nocommandblocks]{progreg}
\usepackage[bracketinterpret,prefixflatinterpret,bracketmodalinterpret,fixformat,differentialdL]{dL}
\usepackage[normalem]{ulem}
\usepackage{proof}
\usepackage{paralist}
\usepackage{tikz}
\usetikzlibrary{arrows}
\usetikzlibrary{calc}
\usetikzlibrary{fit}
\usetikzlibrary{positioning,shadows}
\usetikzlibrary{automata}
\usetikzlibrary{shapes,arrows}
\usetikzlibrary{decorations.text}
\usetikzlibrary{decorations.markings}
\usetikzlibrary{trees,snakes}
\usepackage[strings]{underscore}
\usepackage{textcomp}

\usepackage{prettyref}
\newcommand{\rref}[2][]{\prettyref{#2}}
\newrefformat{sec}{Sect.\,\ref{#1}}
\newrefformat{app}{Appendix\,\ref{#1}}
\newrefformat{def}{Def.\,\ref{#1}}
\newrefformat{thm}{Theorem\,\ref{#1}}
\newrefformat{prop}{Proposition\,\ref{#1}}
\newrefformat{lem}{Lemma\,\ref{#1}}
\newrefformat{cor}{Corollary\,\ref{#1}}
\newrefformat{ex}{Example\,\ref{#1}}
\newrefformat{eq}{Eq.\,\ref{#1}}
\newrefformat{tab}{Table\,\ref{#1}}
\newrefformat{fig}{\figurename\,\ref{#1}}
\newrefformat{case}{case\,\textit{\ref{#1}}}

\newcommand{\VeriPhy}{VeriPhy\xspace}
\newtheorem{example}{Example}
\newtheorem{definition}{Definition}
\newtheorem{theorem}{Theorem}

\begin{document}
\title{A Formal Safety Net for\\Waypoint Following in Ground Robots}

\author{
Brandon Bohrer$^{1,2}$,
Yong Kiam Tan$^{1}$,
Stefan Mitsch$^{1}$,
Andrew Sogokon$^{1}$, and
Andr\'e Platzer$^{1,2}$%
\thanks{Manuscript received: January, 21, 2019; Revised April, 23, 2019; Accepted May, 29, 2019.}
\thanks{This paper was recommended for publication by Editor Dezhen Song upon evaluation of the Associate Editor and Reviewers' comments.
This work was supported by NDSEG, A*STAR, AFOSR grant FA9550-16-1-0288, NSF CPS Award CNS-1739629, the United States Air Force and DARPA under Contract No.\ FA8750-18-C-0092, and the Alexander von Humboldt Foundation.
The views and conclusions contained in this document are those of the authors and should not be interpreted as representing the official policies, either expressed or implied, of any sponsoring institution, the U.S. government or any other entity.}
\thanks{$^{1}$The authors are with Computer Science Department, Carnegie Mellon University, Pittsburgh, USA.
{\tt\footnotesize \{bbohrer,yongkiat,smitsch,asogokon,}
{\tt\footnotesize aplatzer\}@cs.cmu.edu}}%
\thanks{$^{2}$Part of this work was performed while the first and last authors were at \text{Fakult\"at f\"ur Informatik}, \text{Technische Universit\"at M\"unchen.}}%
\thanks{Digital Object Identifier (DOI): \href{https://doi.org/10.1109/LRA.2019.2923099}{10.1109/LRA.2019.2923099}.}
\thanks{\textcopyright 2019 IEEE. Personal use of this material is permitted.  Permission from IEEE must be obtained for all other uses, in any current or future media, including reprinting/republishing this material for advertising or promotional purposes, creating new collective works, for resale or redistribution to servers or lists, or reuse of any copyrighted component of this work in other works.}
}

\maketitle
\begin{abstract}
We present a reusable formally verified safety net that provides end-to-end safety and liveness guarantees for 2D waypoint-following of Dubins-type ground robots with tolerances and acceleration.
We:
    \textit{i)}   Model a robot in differential dynamic logic (\dL), and specify assumptions on the controller and robot kinematics,
    \textit{ii)}  Prove formal safety and liveness properties for waypoint-following with speed limits,
    \textit{iii)} Synthesize a monitor, which is automatically proven to enforce model compliance at runtime, and
    \textit{iv)}  Our use of the VeriPhy toolchain makes these guarantees carry over down to the level of machine code with untrusted controllers, environments, and plans.
The guarantees for the safety net apply to any robot as long as the waypoints are chosen safely and the physical assumptions in its model hold.
Experiments show these assumptions hold in practice, with an inherent trade-off between compliance and performance.
\end{abstract}


\IEEEpeerreviewmaketitle

\newcommand{\admiss}{\textsf{Go}}
\newcommand{\planreq}{\textsf{Feas}}
\newcommand{\veps}{\varepsilon}
\newcommand{\annul}{\textsf{Ann}\xspace}
\newcommand{\adjustSpeedDist}{\delta_\mathsf{Lim}\xspace}
\newcommand{\controllableGoalDist}{\mathsf{Lim}}
\newcommand{\setfml}{J\xspace}
\newcommand{\targetfml}{T\xspace}
\newcommand{\xgivar}{\textsf{xi}}
\newcommand{\ygivar}{\textsf{yi}}
\newcommand{\xgvar}{\textsf{x}}
\newcommand{\ygvar}{\textsf{y}}
\newcommand{\xcvar}{\textsf{xc}}
\newcommand{\ycvar}{\textsf{yc}}
\newcommand{\ytvar}{\textsf{yt}}
\newcommand{\yvar}{\textsf{y}}
\newcommand{\vvar}{\textsf{v}}
\newcommand{\wvar}{\textsf{w}}
\newcommand{\dxvar}{\textsf{dx}}
\newcommand{\dyvar}{\textsf{dy}}
\newcommand{\dxivar}{\textsf{dxi}}
\newcommand{\dyivar}{\textsf{dyi}}
\newcommand{\dxgvar}{\textsf{dxg}}
\newcommand{\dygvar}{\textsf{dyg}}
\newcommand{\kvar}{{\textsf{k}}}
\newcommand{\tvar}{\textsf{t}}
\newcommand{\Avar}{\textsf{A}}
\newcommand{\Bvar}{\textsf{B}}
\newcommand{\Tvar}{\textsf{T}}
\newcommand{\avar}{\textsf{a}}
\newcommand{\vivar}{\textsf{vi}}
\newcommand{\vlvar}{\textsf{vl}}
\newcommand{\vhvar}{\textsf{vh}}
\newcommand{\exctrl}{\textsf{1Dctrl}\xspace}
\newcommand{\explant}{\textsf{1Dplant}\xspace}
\newcommand{\pdrive}{\textsf{go}\xspace}
\newcommand{\pstop}{\textsf{stop}\xspace}
\newcommand{\ctrl}{\textsf{ctrl}\xspace}
\newcommand{\ctrlliv}{\ctrl_{\text{a}}}
\newcommand{\plant}{\textsf{plant}\xspace}
\newcommand{\psimp}{\ensuremath{\alpha_{\textsf{1D}}}\xspace}
\newcommand{\lnorm}[1]{{{\norm{#1}}_{\infty}}}
\newcommand{\enorm}[1]{\norm{#1}}
\newcommand{\ModelPlex}{ModelPlex\xspace}
\newcommand{\linv}{J}

\section{Introduction}
\IEEEPARstart{M}{any} autonomous ground robots are safety-critical because they operate near or in concert with humans.
Formally verifying these systems is important: logic allows rigorous correctness arguments that apply in \emph{all} system states, providing a powerful complement to system testing.
Yet for robotics, even choosing a property to formally verify is challenging: many modeling abstractions and safety properties are available, with competing trade-offs.
Discrete techniques can be applied to control software, but robots are \emph{cyber-physical systems}: their verification must account for discrete controllers, continuous mechanics, and interactions between them.

Hybrid systems~\cite{DBLP:conf/lics/Henzinger96} emerged as mathematical models for robots because they integrate discrete with continuous dynamics, but they raise new questions about what it means to verify a robot.
Robot kinematics are endlessly nuanced, so any model is always an approximation of reality.
Even control software is rarely modeled exactly: simplifications are necessary in practice to limit verification complexity.
Moreover, control software evolves throughout its development, and since verification of arbitrary programs cannot be fully automated,
re-verifying control code after every change would be impractical.

How then ought a robot be verified?
Online monitoring, per the Simplex method~\cite{Krogh1998TheSA}, offers a solution: run the control software, but treat its control decision as an \emph{untrusted suggestion} which is supervised against a trusted \emph{monitor condition} $\phi$ describing all ``known-safe'' decisions.
Whenever the suggestion is known-safe the supervising \emph{monitor} allows it to proceed, but otherwise invokes a trusted fallback decision, like emergency braking, to regain safety before returning control to the untrusted controller.
Online monitoring is appealing because it enables treating the controller as a black box: only the monitor condition and fallback are safety-critical, both of which are simpler and can often be re-used as the control software evolves, or even across hardware platforms.
To promote reusability, we target a waypoint-following notion of safety: other notions of safety such as collision-avoidance often come with restrictive assumptions, e.g., on the quantity and dynamics of obstacles, whereas waypoint-following abstracts away such problems under the choice of safe waypoints.

The monitor conditions and fallback are both best kept simple. 
While steering enables active safety and may reduce the required braking power, we simply brake at the maximum rate.
Regardless which fallback is used, however, what is essential is that the monitor conditions and fallback provide safety.
The \ModelPlex~\cite{DBLP:journals/fmsd/MitschP16} synthesizer and \VeriPhy~\cite{DBLP:conf/pldi/BohrerTMMP18} compilation toolchain ensures safety at implementation level by:
\begin{inparaenum}[\it i)]
  \item synthesizing correct-by-construction \emph{monitor conditions} $\phi$ from a proven-safe hybrid systems model containing a proven-safe fallback (\ModelPlex)
  \item soundly compiling high-level monitor conditions and high-level fallback programs to machine-code monitor implementations with sound machine arithmetic (\VeriPhy).
\end{inparaenum}
\ModelPlex and \VeriPhy expand upon the reusability inherent to the black-box approach: hybrid systems models and proofs can treat many system parameters (e.g., tolerances and system delay) \emph{generically}, once and for all, for all choices of the parameters.
The hybrid system model can be used as a template: new monitors can often be generated for new systems \emph{without} doing new proofs, so long as choosing system parameters suffices to faithfully model the new system.
The \VeriPhy approach is also \emph{end-to-end} in that \VeriPhy outputs a chain of formal proofs, in theorem provers, that the actions taken by the machine-level (monitored) program fall under the original model and are thus formally safe.
Even then, some guarantees are lost when sensors and actuators are buggy, when monitor conditions for the physical dynamics are violated, or when an unsafe plan is provided: we discuss these limitations in related work~(\rref{sec:related-work}).

The \VeriPhy approach has until now only been tested on an overly simplistic low-speed robot implementing a model of straight-line motion with direct velocity control.
In this paper, we show for the first time that this approach scales to realistic models and simulations:
\begin{itemize}
\item We model as a hybrid program~\cite{DBLP:books/sp/Platzer18} a 2D accelerating vehicle that follows bloated Dubins paths.
\item We prove safety and liveness properties for the model with the \KeYmaeraX theorem prover~\cite{DBLP:conf/cade/FultonMQVP15} for differential dynamic logic (\dL)~\cite{DBLP:books/sp/Platzer18}.
\item We use \VeriPhy~\cite{DBLP:conf/pldi/BohrerTMMP18} to synthesize a monitor and to automatically prove its correctness down to the machine code implementation, which gives an end-to-end proof.
\end{itemize}

Waypoint-following has the advantage of a clean interface to other robot software with its dual purposes of both \emph{safety} (avoiding unsafe regions) and \emph{liveness} (reaching its goal).
We call waypoint-following safe iff the robot always follows the given path to its waypoint within a given tolerance and obeys given speed limits.
Collision freedom then reduces to checking that correct waypoints and speed limits were given.
The system is \emph{live} if at all points it is possible to drive the rest of the way to the waypoint.

Obstacle avoidance \cite{DBLP:journals/ijrr/MitschGVP17}, in contrast, directly verifies collision freedom, but liveness is challenging to even state let alone prove.
The mix of safety and liveness is essential because a motionless robot is technically safe, but neither live nor useful because it never reaches its goal.
By studying safety and liveness of waypoint-following, we provide a clean separation of concerns compared to the orthogonal question of verifying a discrete planner~\cite{DBLP:conf/atva/RizaldiISA18}.

Bloating the 2D Dubins dynamics adds an additional tolerance margin to the ideal dynamics which accounts for the gap between the approximate dynamical model and reality (\rref{sec:robot-model}).
Our evaluation shows that:
\begin{inparaenum}[\it i)]
\item a variety of classical control choices such as bang-bang and PD control fit within the bloated ideal path
\item the model assumptions hold in practice because AirSim's non-holonomic dynamics fall within bloated ideal holonomic Dubins dynamics, and
\item there is a trade-off between meeting model assumptions and operational performance: more aggressive controllers break assumptions more often.
\end{inparaenum}
This paper also serves as a case study on safety and liveness verification: once the \emph{safety} property is proved, much of the effort can be reused to prove \emph{liveness}.
The safety and liveness proof were performed interactively, but crucially need only be performed once per \dL model, which describes an entire class of systems.
Thanks to the automated proofs provided by \VeriPhy, runtime monitors can be applied to new controls and even new hardware or simulation platforms with no additional manual proofs, so long as the controls and dynamics stay within a tolerance around the ideal holonomic dynamics, so that the same \dL model applies.
Because the model treats system parameters (such as system delays and tolerances) generically, a wide variety of Dubins-like systems are already supported
simply by changing the parameters.
That is how we developed a formal safety net for ground robots and evaluated it on a realistic simulation.
Because the monitor is reusable, we hope our safety net can assist future implementers in developing new systems.

\section{Related work}
\label{sec:related-work}
Related work in formal methods and robotics applied synthesis and verification techniques to safe robotic control.
This paper is the first to use a verified-safe monitor to enforce waypoint-following correctness of a realistic simulation.

\subsection{Synthesis for Verified Planning and Control}
Much of the existing related work considers high-level plan synthesis in isolation, with informal proofs of correctness.
Our work is complementary: we address correctness of low-level control, provide formal guarantees, and \emph{check} rather than \emph{assume} that runtime physics matches the model:
\begin{itemize}
\item The tools LTLMoP~\cite{DBLP:conf/iros/FinucaneJK10} and TuLiP~\cite{DBLP:conf/IEEEcca/FilippidisDLOM16} synthesize robot controls that satisfy a temporal logic specification.
They excel at providing an intuitive user interface for specifying discrete planning problems, though discretization~\cite{DBLP:journals/tac/LiuOTM13,DBLP:journals/automatica/FainekosGKP09} can be used to support continuous dynamics.
We focus instead on providing the highest degree of confidence by proving safety in a theorem prover, including proofs of the dynamics and down to machine-code level.
\item Controllers have been synthesized:
\begin{inparaenum}[\it i)]
\item from temporal logic specifications for linear systems~\cite{DBLP:journals/tac/KloetzerB08},
\item for adaptive cruise control~\cite{DBLP:journals/tcst/NilssonHBCAGOPT16}, tested in simulation and on hardware, and
\item from safety proofs~\cite{DBLP:conf/emsoft/TalyT10} for switched systems using templates.
\end{inparaenum}
These all assume model compliance and cannot ensure feedback controller correctness.
\end{itemize}

\subsection{Offline Verification for Planning and Control}
In contrast to online synthesis, offline verification can show safety in all uncountably many states.
High-level models of the system under consideration can already be verified during the design phase of a project when changes are cheap.
Much robotics verification work focuses on hybrid systems models;
common approaches are reachability analysis~\cite{DBLP:conf/cav/FrehseGDCRLRGDM11} and theorem proving~\cite{DBLP:conf/cade/FultonMQVP15}.
Both have been applied in case studies~\cite{DBLP:journals/ijrr/MitschGVP17,TomlinMitchell} and experience shows that reachability typically provides more automation while theorem proving supports a powerful combination of rigorous foundations and establishes guarantees for unbounded time and space.
Both approaches can be combined with monitoring:
\begin{itemize}
\item 1D straight-line motion was addressed both in \dL~\cite{DBLP:conf/pldi/BohrerTMMP18} and with reachability analysis~\cite{chen2015benchmark},
but 1D uses simpler verification technology and is not suitable for real robots.
\item
 Unbounded-time 2D obstacle avoidance and 1D liveness have also been proved in \dL~\cite{DBLP:journals/ijrr/MitschGVP17}, and
 liveness has been proved on paper~\cite{DBLP:journals/corr/abs-1709-02561}.
 Their controllers, like ours, are related to the Dynamic-Window~\cite{DBLP:journals/ram/FoxBT97} algorithm.
 Our novel results include 2D liveness, waypoint-following, and end-to-end correctness.
 While collision avoidance is simpler in prior work, their approach precludes liveness, which we proved.
 Prior \dL efforts treat sensor errors explicitly, for which synthesis is subject of ongoing work~\cite{DBLP:journals/corr/abs-1811-06502}.
 In contrast, we integrate synthesized monitors with a simulation while keeping guarantees.
 To this end, we use a single tolerance for sensing/actuation error and deviation of real dynamics from the model.
 The limitation of this approach is that our guarantees do not explicitly incorporate sensor errors.
\item
  A planner for ground vehicles was verified~\cite{DBLP:conf/atva/RizaldiISA18} in Isabelle.
  Their physics are close to ours, but feedback control and implementation correctness are not addressed.

\end{itemize}

\subsection{Online Verification}
Online/runtime verification provides a runtime safety net, but the correctness of the safety net itself is then critical to system safety.
In contrast to offline verification, online verification cannot predict safety for infinitely many states.
\begin{itemize}
\item The basis of online verification is the Simplex~\cite{Krogh1998TheSA} method, which uses a trusted monitor to decide between an untrusted controller and trusted fallback.
\item The \VeriPhy~\cite{DBLP:conf/pldi/BohrerTMMP18} toolchain for \dL, which we use, combines offline and online verification to extend Simplex by ensuring the monitor is correct-by-construction, formally proving its safety, and maintaining those guarantees down to machine code implementations.
\item
Runtime monitoring has been combined with nonexhaustive model checking and evaluated in simulation~\cite{DBLP:conf/rv/DesaiDS17}.
Their relative strengths are in correctness of high-level event-handling logic and experimentally learning tolerances for the dynamics.
Our relative strengths are use of a theorem-prover to show safety in \emph{all} states, richer physical dynamics, and correct-by-construction monitors.
\item Runtime reachability analysis has been used for Dubins-like car control~\cite{DBLP:journals/trob/AlthoffD14},
but runtime model compliance is not enforced and the reachability checker is trusted.
\end{itemize}

\subsection{Simulation}
Simulation is an essential part of evaluating models and designs.
We used the AirSim~\cite{DBLP:conf/fsr/ShahDLK17} simulator for autonomous cars (originally for UAVs), because it comes with accurate physical and visual models out-of-the-box.
Using these existing models provides a degree of independence in our evaluation.

In short, while verification of robotics receives frequent attention, few works have addressed rigorous end-to-end guarantees.
We develop the first realistic system with formal end-to-end safety and liveness guarantees for 2D waypoint following, by generating a runtime monitor from a verified model.
Crucially, we expect this runtime safety net can be applied to other Dubins-like system without redoing any proofs.

\section{Background: Differential dynamic logic}
\newcommand{\ivr}{\psi\xspace}
We write our model as a \emph{hybrid program} and use \emph{differential dynamic logic} (\dL)~\cite{DBLP:books/sp/Platzer18} to verify it.
Hybrid programs express \emph{hybrid systems} as programs containing differential equations (ODEs).
They are particularly useful for verified robotics because they concisely describe both the control laws and kinematics of the system.
\rref{tab:hybrid-programs} gives the syntax of hybrid programs and informally describes their semantics, wherein running a program $\alpha$ results in zero, one, or many different states.
Detailed formal semantics are provided elsewhere~\cite{DBLP:books/sp/Platzer18}.
\begin{table}[htbp!]
\centering
\caption{Hybrid programs}
\label{tab:hybrid-programs}
\begin{tabular}{ll}
    Program & Means   \\\hline
$\ptest{\phi}$              & Results in current state if $\phi$ is true, no states if false.\\
$\humod{x}{\theta}$    & Store value of expression $\theta$ in variable $x$.\\
$\prandom{x}$            & Store arbitrary (real) number in variable $x$.\\
$\pevolvein{\D{x}=\theta}{\ivr}$ & Evolve ODE $\D{x} = \theta$ for any duration $t{\geq}0,$ \\
& with constraint formula $\ivr$ true throughout.\\
$\alpha;\beta$         & Run $\alpha,$ then $\beta$ in any resulting state(s).\\
$\alpha\cup\beta$      & Choose between running $\alpha$ or $\beta$.\\
$\alpha^*$             & Repeats $\alpha$ $n$ times, for any $n\in\mathbb{N}$.
\end{tabular}
\end{table}
Typical controllers use assignments $\humod{x}{\theta}$ to store the value of (polynomial) expression $\theta$ in variable $x,$ or assign an arbitrary value ($\prandom{x}$) and then test ($\ptest{\phi}$) that the value satisifies some condition $\phi$.
Choice ($\alpha\cup\beta$) allows choosing between control laws, each of which may have (overlapping) tests ($\ptest{\phi}$) saying when each law applies.
Semicolons separate statements, so sequencing $(\alpha;\beta)$ runs $\beta$ after $\alpha$, while loop $\prepeat{\alpha}$ repeats $\alpha$ any arbitrary number of times.
Many models follow the \emph{control-plant loop} idiom (e.g., $\psimp \equiv \prepeat{(\exctrl;\explant)}$),
where a discrete program $\exctrl$ is followed by a continuous $\explant$ modeling physics, repeated in a loop (*).
The $\explant$ is an ODE $\pevolvein{\D{x}=\theta}{\ivr},$ which evolves according to $\D{x}=\theta$ for any duration such that $\ivr$ holds throughout.
Before we develop a realistic 2D model in \rref{sec:robot-model},
we recall, in \rref{ex:straight-vel}, a toy example, \psimp, of 1D motion with perfect speed control~\cite{DBLP:conf/pldi/BohrerTMMP18}.
The controller can either $\pdrive$ forward with some $v$ such that ${0}{\leq}{v}{\leq}{V}$ if we are far enough ($d\geq TV$) from the destination $d$, else it must $\pstop$ by setting $v$ to 0.
The differential equation $\D{d}=-v, \D{t}=1$ says the distance $d$ continuously decreases proportional to velocity $v,$ while time continuously elapses at rate 1.
The constraint $t \leq T$ after $\&$ is a \emph{time trigger}, saying that at most $T$ seconds may elapse between control cycles.
Note that we will show safety for any number of control cycles, and thus for \emph{unbounded time}.
\begin{example}[Simple 1D Idealized Driving]
\label{ex:straight-vel}
{\small  \begin{align*}
\exctrl  &\equiv \pdrive\cup\pstop \!\!\!\!\!\!\!\!\!\!\!\!\!\!& \pdrive  &\equiv~\ptest{d{\geq}TV};~\prandom{v};~ ?{0}{\leq}{v}{\leq}{V}\\
\pstop  &\equiv \humod{v}{0}       & \explant &\equiv~\humod{t}{0};~\{\pevolvein{\D{d}=-v, \D{t}=1}{t \leq T}\}
  \end{align*}}
\end{example}
\noindent
Formulas of \dL are used to formalize program properties:
\begin{definition}[\dL formulas]
Formulas $\phi,\psi$ of \dL consist of the following connectives:
\begin{align*}
\phi,\psi ::= \phi\land\psi\ &|\ \phi\lor\psi\ |\ \phi\limply\psi\ |\ \neg\phi\ |\ \theta_1 \sim \theta_2\\
          |\ \forall{x}{~\phi}\ &|\ \exists{x}{~\phi}\ |\ \dbox{\alpha}{\phi}\ |\ \ddiamond{\alpha}{\phi}
\end{align*}
\end{definition}
\noindent
where $\phi\land\psi$ holds when $\phi$ and $\psi$ both hold, $\phi\lor\psi$ holds when either $\phi$ or $\psi$ holds, $\phi\limply\psi$ holds if $\psi$ holds assuming $\phi$ holds, $\neg\phi$ holds if $\phi$ does not, and $\forall{x}{~\phi}$ and $\exists{x}{~\phi}$ hold if $\phi$ holds for all or some value(s) of $x$, respectively.
When we prove a theorem $\phi,$ all variables implicitly have a ``for all'' quantifier so, e.g., the safety and liveness theorems hold for \emph{all} states and \emph{all} values of system parameters.
Formula $\theta_1 \sim \theta_2$ is shorthand for any comparison $\sim\mathop{\in}\{\leq,<,=,\neq,>,\geq\}$ where $\theta_i$ are real multivariate polynomials.
The modalities $\dbox{\alpha}{\phi}$ and $\ddiamond{\alpha}{\phi}$ say $\phi$ holds in \emph{all} or \emph{some} state(s) reached by executing hybrid program $\alpha$ respectively; they are used to express safety and liveness properties for our models.

\rref{eq:straight-safety} is a safety formula for $\psimp$:
If the robot has not collided initially ($d \geq 0$) then the verified model will never collide no matter how many further control cycles are executed:
\begin{equation}
\label{eq:straight-safety}
d \geq 0 \land V \geq 0 \land T \geq 0  \limply \dbox{\prepeat{(\exctrl;\explant)}}{\,d \geq 0}
\end{equation}
This toy example, which was previously used to demonstrate \VeriPhy~\cite{DBLP:conf/pldi/BohrerTMMP18}, misses out on many of the challenges essential to robotics: curved motion, acceleration, actuation disturbance, and goal-following all demand more sophisticated control conditions and invariants, which demand more sophisticated proof techniques.
We take on these challenges in \rref{sec:robot-model}.
\section{Ground Robot Model}
\label{sec:robot-model}
This section introduces our 2D robot model in \dL.
This model is the heart of our verification effort: it lays out the definition of safety, assumptions on the controller, and assumptions on the plant.
It will enable us in \rref{sec:ver-reachavoid} to prove that these assumptions are strong enough to guarantee safety, then in \rref{sec:validation} to synthesize a monitor which functions as a runtime safety net, providing formal safety guarantees.
The liveness proof of \rref{sec:ver-reachavoid} complements safety and increases confidence in the model by showing our model is never so restrictive that it would force the robot to get stuck.

We use waypoint-following because it covers a wide variety of realistic scenarios, whereas collision avoidance is challenging to specify formally without making overly restrictive assumptions~\cite{DBLP:journals/ijrr/MitschGVP17}.
The implementation trusts waypoints from a planner.
Feedback control is considered safe so long as it follows the waypoints within a fixed desired tolerance.
The tolerance accounts for imperfect actuation and for discrepancies between the ideal dynamics and real dynamics of the implementation.
The tolerance may also be increased to account for bounded sensor error, but the formal guarantees provided here are for perfect sensing.

Each waypoint is specified by coordinates $(\xgvar,\ygvar)$, a curvature $\kvar$, and a speed limit $[\vlvar,\vhvar]$ for the robot's velocity $\vvar$.
By convention, positive $\xgvar$ points \emph{forward}, and positive $\ygvar$ points \emph{left}.
The curvature $\kvar$ yields circular arcs (when $\kvar \neq 0$) and lines (when $\kvar=0$) as primitives.
The addition of speed limits allows a plan to specify, for example, that the robot ought to slow down for a sharp curve or stop.
The speed limits need only be met at the endpoint of the waypoints, which improves monitor compliance (in \rref{sec:validation}).
Because realistic robots never follow a path perfectly, we bloat each arc to an annular section which is more easily followed.
Our hybrid program $\alpha$ is again a time-triggered \emph{control-plant} loop: $\alpha\equiv(\ctrl;\plant)^*$.

\begin{figure}[h!tb]
\centering
\includegraphics[width=0.3\textwidth]{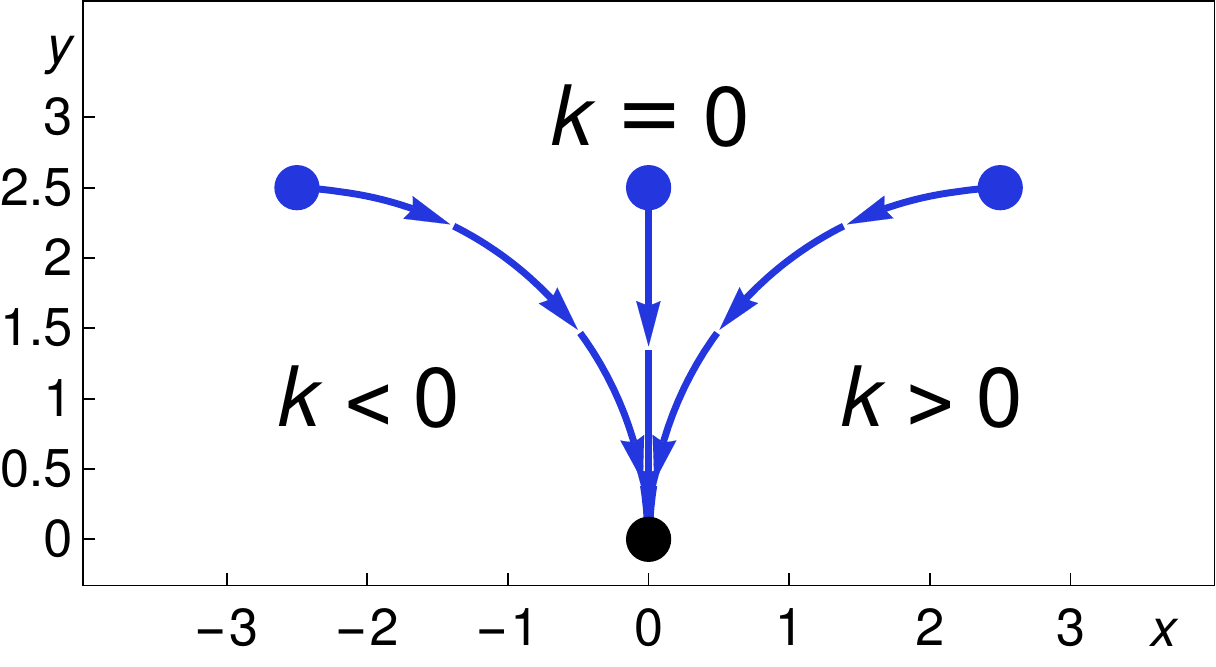}
\caption{Trajectories of dynamics for different choices of \kvar.}
\label{fig:ode1}
\end{figure}

We use \emph{relative} coordinates: the robot's position is always the origin, from which perspective the waypoint ``moves toward'' the vehicle.
This simplifies proofs (fewer variables) and implementation (real sensors and actuators are vehicle-centric).
\rref{fig:ode1} illustrates control scenarios for the system in relative coordinates.
The robot is represented by \includegraphics[height=1em]{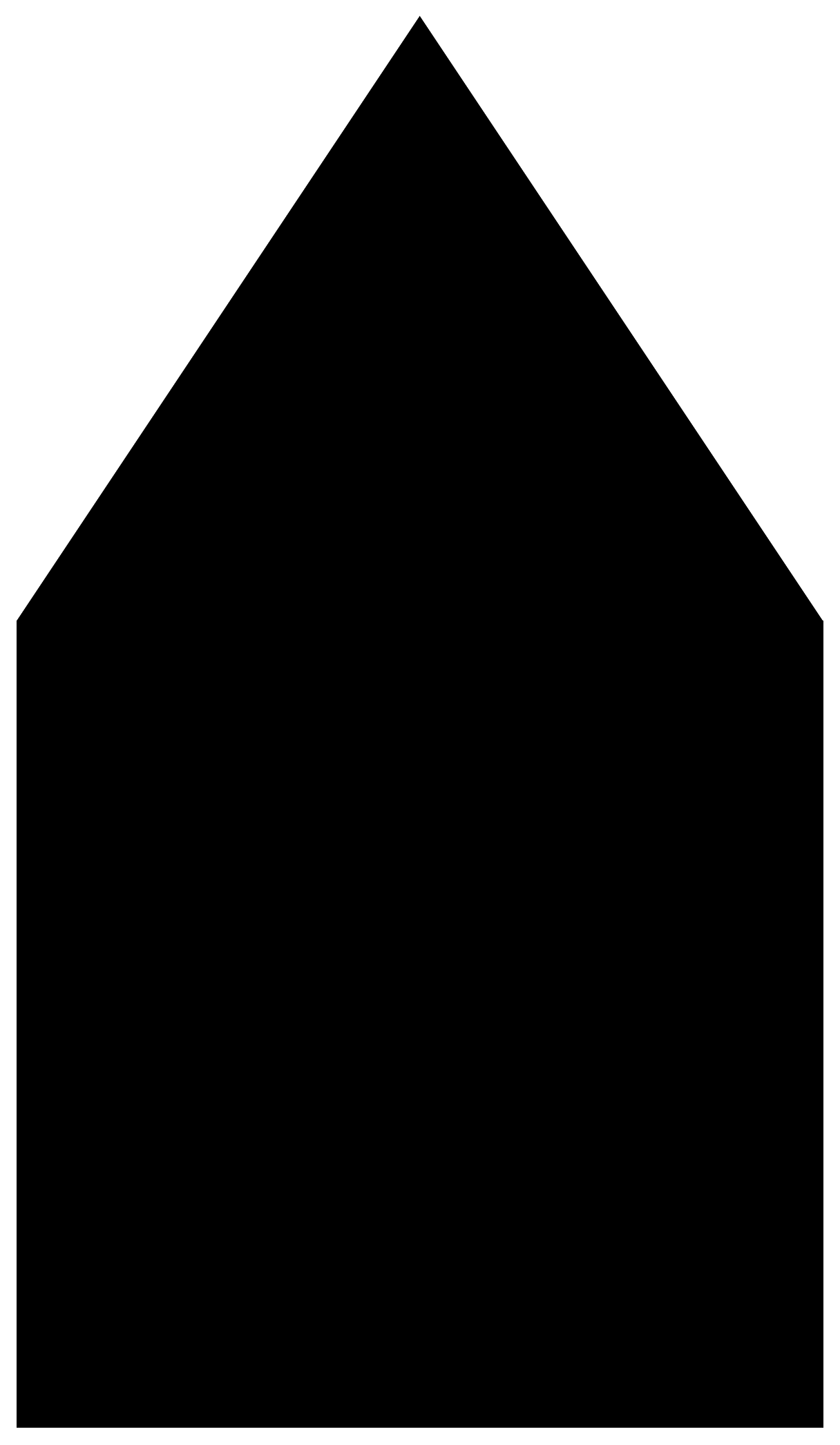} where the triangle points in the robot's forward direction.
The control choice $\kvar = 0$ drives waypoints that are straight ahead of the robot straight towards the robot.
For waypoints initially to the left of the robot, a control choice $\kvar > 0$ yields clockwise motion of the waypoint towards the robot.
Conversely, for waypoints initially to the right, control choices with $\kvar < 0$ yield counter-clockwise motion.

The relative coordinate system and control choices for $\kvar$ are modeled by the ODE $\plant$:
\begin{align*}
\plant\equiv \humod{\tvar}{0};~\{&\D{\xgvar}=\vvar~(\kvar~\ygvar-1),~\D{\ygvar}=-\vvar~\kvar~\xgvar,~\D{\vvar}=\avar,\\
 &\D{\tvar}=1~\&~\tvar\leq \Tvar ~\land~ \vvar \geq 0\} 
\end{align*}
Here, $\avar$ is an input from the controller describing the acceleration with which the robot is to follow the arc of curvature $\kvar$ to waypoint $(\xgvar,\ygvar)$.
In the equations for $\D{\xgvar},\D{\ygvar}$: \begin{inparaenum}[\it i)]\item The $\vvar$ factor models $(\xgvar,\ygvar)$ moving at speed $\vvar$, \item The $\kvar,\xgvar,\ygvar$\ factors model circular motion with curvature $\kvar$.
\item The additional $-1$ term in the $\D{\xgvar}$ equation shifts the center of rotation to $\left(0,\frac{1}{\kvar}\right)$.
\end{inparaenum}
The equations $\D{\vvar}=\avar$ and $\D{\tvar}=1$ make acceleration the derivative of velocity and $\tvar$ stand for current time.
The \emph{domain constraint} $\tvar \leq \Tvar \land \vvar \geq 0$ after $\&$ says that the duration of one control cycle shall never exceed the timestep parameter $\Tvar > 0$ representing the maximum delay between control cycles and that the robot never drives in reverse.

\rref{fig:ode1} depicts curves that are exact solutions of $\plant$ where the robot exactly meets the waypoint.
Because realistic robots cannot follow these curves exactly, the waypoint is bloated by a fixed ball of radius $\varepsilon > 0,$ giving the robot some freedom in curve-following. We refer to this bloated waypoint as the \emph{goal} for the robot.
\rref{fig:ode2} illustrates a goal of size $\varepsilon=1$ around the origin and several trajectories which pass through the goal.

{\begin{figure}[h!t]
\centering
\includegraphics[width=0.3\textwidth]{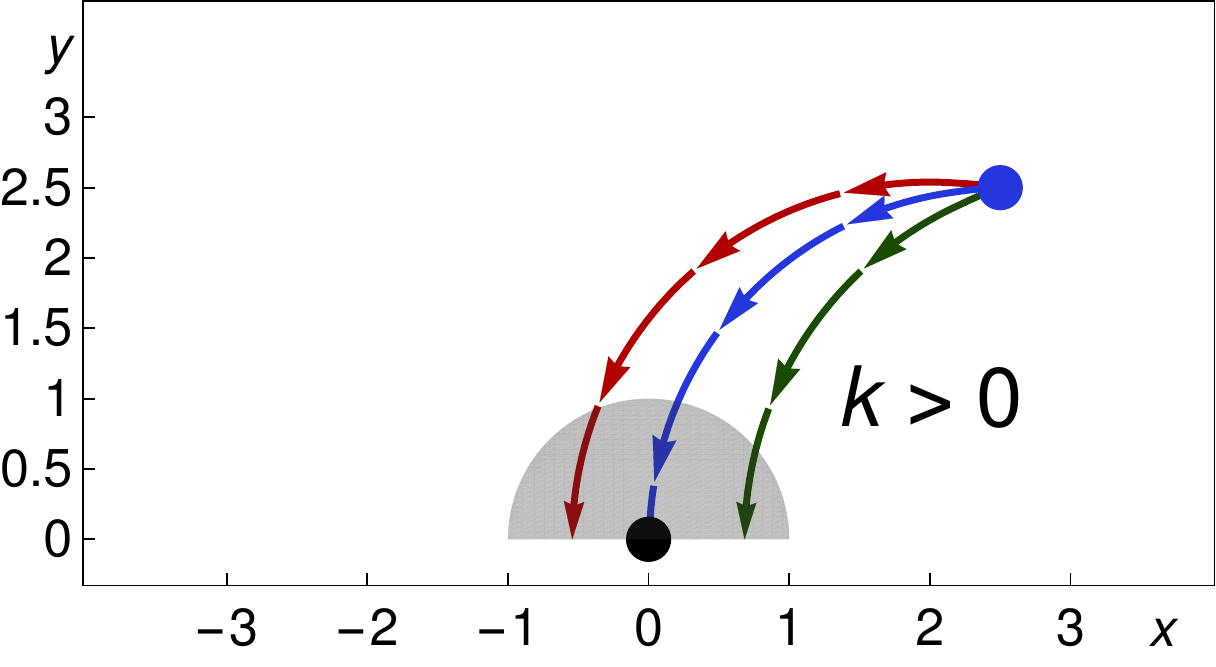}
\caption{Trajectories of \plant for choices of $\kvar < 0$ when $\varepsilon = 1$.}
\label{fig:ode2}
\end{figure}}

The controller's task is to compute an acceleration $\avar$ which slows down (or speeds up) soon enough that the speed limit $\vvar \in [\vlvar,\vhvar]$ is ensured \emph{by the time} the robot reaches the goal.
We allow the controller to exceed $\vhvar$ temporarily as long as there is time to achieve $\vvar \leq \vhvar$ before reaching the goal.
This relaxation improves monitor compliance and lets the robot speed up more quickly, e.g., when it is far from the goal.
The trade-off is that the proof becomes more challenging than it would be for a model that always enforces limits.
The controller is written:

{{\small\begin{align*}
\ctrl &~\equiv~ \prandom{(\xgvar,\ygvar)};\,\prandom{[\vlvar,\vhvar]};\,\prandom{\kvar};\,\ptest{\planreq};\,\underbrace{\prandom{\avar};\,\ptest{\admiss}}_{\ctrlliv}\\
\annul &~\mequiv~\abs{\kvar}\veps \leq 1 \land \Big|\frac{ \kvar~\left(\xgvar^2 +\ygvar^2 - \veps^2 \right)}{2} - \ygvar\Big| < \veps\\
\label{eq:planReq} \planreq &~\mequiv~ \annul \land \xgvar{>}0 \land 0{\leq}\vlvar{<}\vhvar \land \Avar\Tvar{\leq}\vhvar-\vlvar \land \Bvar\Tvar{\leq}\vhvar-\vlvar\\
      \admiss& ~\mequiv~ {-}\Bvar\leq \avar \leq\Avar \land \vvar + \avar\Tvar \geq 0 \\
\land & \Bigl(\vvar \leq \vhvar \land \vvar+\avar\Tvar\leq \vhvar ~\lor \\
      & (1+\abs{\kvar}\veps)^2 \Bigl(\vvar\Tvar+\frac{\avar}{2}\Tvar^2 + \frac{(\vvar{+}\avar\Tvar)^2-\vhvar^2}{2\Bvar}\Bigr) + \veps{\leq}\lnorm{(\xgvar,\ygvar)} \Bigr)\\
\land & \Bigl(\vlvar \leq \vvar \land \vlvar \leq \vvar+\avar\Tvar ~\lor\\
      & (1+\abs{\kvar}\veps)^2 \Bigl(\vvar\Tvar+\frac{\avar}{2}\Tvar^2 + \frac{\vlvar^2 - (\vvar{+}\avar\Tvar)^2}{2\Avar}\Bigr) + \veps{\leq}\lnorm{(\xgvar,\ygvar)}\Bigr)\end{align*}}}
\noindent where the nondeterministic assignment $\prandom{(\xgvar,\ygvar)}$ chooses the next 2D waypoint, the assignment $\prandom{[\vlvar,\vhvar]}$ chooses the speed limit interval, and $\prandom{\kvar}$ chooses any curvature.
The subsequent \emph{feasibility} test $\ptest{\planreq}$ checks whether or not the chosen waypoint, speed limit, and curvature are \emph{physically} feasible in the current state under the $\plant$ dynamics (e.g., that there is enough remaining distance to get within the speed limit interval).
We also simplify plans so that all waypoints satisfy $\xgvar > 0$ by subdividing any violating paths automatically.
This simplifies the feedback controller and proofs.

In $\planreq,$ formula $\annul$ says we are within the \emph{annular section} (\rref{fig:circlestaging}) ending at the waypoint $(\xgvar, \ygvar)$ with the specified curvature $\kvar$ and width $\veps$.
A larger choice of $\veps$ yields more error tolerance in the sensed position and followed curvature at the cost of an enlarged goal region.
Formula $\annul$ also contains a simplifying assumption that the radius of the annulus is at least $\veps$.
$\planreq$ also says the speed limits are assumed distinct and large enough to not be crossed in one control cycle.

\begin{figure}[h!]
\centering
\includegraphics[width=0.3\textwidth]{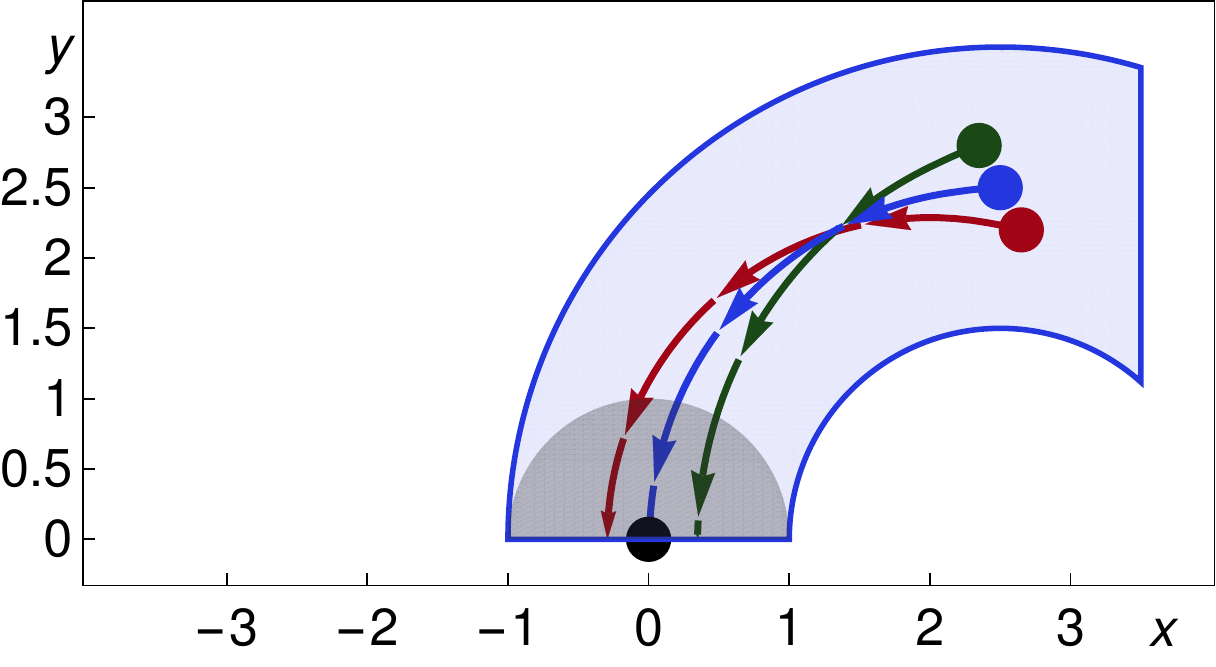}
\caption{Annular section through the (blue) waypoint $(2.5,-3)$. Trajectories from the displaced green and red waypoints with slightly different curvatures remain within the annulus.}\label{fig:circlestaging}
\label{fig:circlestaging}
\end{figure}

The \emph{admissibility} test $\ptest{\admiss}$ checks that the chosen $\avar$ will take the robot to its goal with a safe speed limit, by \emph{predicting future motion} of the robot.
We illustrate this with the upper bound conditions.
The bound will be satisfiable after one cycle if either the chosen acceleration $\avar$ already maintains speed limit bounds ($\vvar \leq \vhvar \land \vvar{+}\avar\Tvar \leq \vhvar$) or when there is enough distance left to restore the limits before reaching the goal.
For straight line motion ($\kvar=0$), the required distance can simply be found by integrating acceleration and speed:
\begin{equation*}
\overbrace{\vphantom{\frac{\Tvar}{\Bvar}} \vvar\Tvar+\frac{\avar}{2}\Tvar^2}^{\mathclap{\text{distance at}~\Tvar}} + \frac{(\overbrace{\vvar{+}\avar\Tvar}^{\mathclap{\text{speed at}~\Tvar}})^2-\vhvar^2}{2\Bvar} + \veps \leq \lnorm{(\xgvar,\ygvar)}
\end{equation*}
where $\avar \in [-\Bvar,\Avar]$.
The extra factor of $(1 + \abs{k}\veps)^2$ for curved motion accounts for the fact that an arc along the inner side of the annulus is shorter than one along the outside (\rref{fig:circlestaging}).

The abbreviation $\ctrlliv$ names just the control code responsible for deciding \emph{how} the waypoint is followed rather than \emph{which} waypoint is followed.
We show in \rref{sec:validation} that $\planreq$ and $\admiss$ are easily satisfied by the implementation.
We verify safety and liveness of the above model in \rref{sec:ver-reachavoid}.
\section{Formal Safety and Liveness Guarantees}
\label{sec:ver-reachavoid}
We now state the safety and liveness theorems in \dL:
\begin{theorem}[Safety]
\label{thm:safe}
The following \dL formula is valid:
\begin{align*}
\Avar{>}0 & \land \Bvar{>}0 \land \Tvar{>}0 \land \veps{>}0 \land \linv \limply \\
&\dbox{\prepeat{(\ctrl;\plant)}}{\big(\enorm{(\xgvar,\ygvar)} \leq \veps \limply \vvar \in [\vlvar,\vhvar]\big)}
\end{align*}
where validity means the formula holds in all states and, e.g., all admissible waypoints.
\end{theorem}

The first four assumptions ($\Avar{>}0 \land \dots \land \veps{>}0$) are basic sign conditions on the symbolic constants used in the model.
The final assumption, $\linv$, is the loop invariant.
The full definition of $\linv$ is deferred to \rref{sec:robostage}, but captures the fact that the robot never strays far from its path.
We write $\enorm{(\xgvar,\ygvar)}$ for the Euclidean norm $\sqrt{\xgvar^2 + \ygvar^2}$ and consider the robot ``close enough'' to the waypoint when $\enorm{(\xgvar,\ygvar)} \leq \veps$ for our chosen goal size $\veps$.
The theorem states that no matter which (admissible) control decisions are made, whenever the vehicle is in the goal region of size $\veps$, it obeys the speed limit $\vvar \in [\vlvar,\vhvar]$.
While this provides a formal notion of safety, it does not prove that the robot can actually reach the goal, which is a \emph{liveness} property:
\begin{theorem}[Liveness]
\label{thm:liveness}
The following \dL formula is valid:
\begin{align*}
\Avar{>}0 &\land \Bvar{>}0 \land \Tvar{>}0 \land \veps{>}0 \land \linv \limply \\
&\dbox{\prepeat{(\ctrl;\plant)}}{\Big( \vvar{>}0 \land \xgvar{>}0 \limply }\\
&\ \ \ \ \ddiamond{\prepeat{(\ctrlliv;\plant)}}{\big(\enorm{(\xgvar,\ygvar)} \leq \veps~\land~\vvar \in [\vlvar,\vhvar]\big)} \Big)
\end{align*}
\end{theorem}

Under the same assumptions as \rref{thm:safe}, this theorem says that no matter how long the robot has been running ($\dbox{\prepeat{(\ctrl;\plant)}}{}$) already,
if some simplifying assumptions still hold ($\vvar{>}0\land\xgvar{>}0$)
the controller can be run ($\ddiamond{\prepeat{(\ctrlliv;\plant)}}{}$) with admissible acceleration choices $(\ctrlliv)$ to reach the present goal \((\enorm{(\xgvar,\ygvar)} \leq \veps)\) within the desired speed limits \((\vvar \in [\vlvar,\vhvar])\).
The simplifying assumptions $\vvar{>}0\land\xgvar{>}0$ say the robot is still moving forward and the waypoint is still in the upper half-plane, i.e., it has not driven \emph{past} the waypoint.

\subsection{Proving Safety}
\label{sec:robostage}
We give the high-level insights here, such as invariants. We first prove the  program stays in the region where $\linv$ is true, then that $\linv$ implies safety.
To satisfy the safety condition, our program must maintain an invariant $\linv$ that:
\begin{inparaenum}[\it i)]
\item the robot follows the plan closely and
\item it drives at speeds that let it achieve the speed limits in the remaining distance to the goal.
\end{inparaenum}

Predicate $\controllableGoalDist(v_1,v_2,a,\xgvar,\ygvar,\kvar)$ says acceleration $a$ can close velocity gap $v_1 - v_2$ before waypoint $(\xgvar,\ygvar)$ reaches the origin:
\begin{align*}
\adjustSpeedDist(v_1, v_2, a) &\equiv (1+\abs{\kvar}\veps)^2~\frac{v_1^2-v_2^2}{2a}\\
\controllableGoalDist(v_1, v_2, a, \xgvar, \ygvar) &\equiv v_1{\leq}v_2 \lor \adjustSpeedDist(v_1,v_2,a) + \veps {\leq}\lnorm{(\xgvar,\ygvar)}
\end{align*}
Here, $\adjustSpeedDist(v_1, v_2, a)$ bounds distance needed to change velocity from $v_1$ to $v_2$ with a scaling factor $(1+\abs{\kvar}\veps)^2$ for the tolerance incurred from the width of the annular section compared to an arc.
This is not too conservative in practice: it is tightest (near $1$) for small $\veps$ or small $\abs{\kvar}$ and never exceeds $4$.
This suffices to define $\linv$, which says the speed limits can be achieved with maximum braking ($\Bvar$) and acceleration ($\Avar$):
\begin{multline*}
    J \equiv \annul \land 0\leq\vlvar < \vhvar \land \Avar\Tvar \leq \vhvar-\vlvar \land \Bvar\Tvar \leq \vhvar-\vlvar\\
    \land \controllableGoalDist(\vlvar,\vvar, \Avar, \xgvar, \ygvar) \land \controllableGoalDist(\vvar,\vhvar,\Bvar,\xgvar,\ygvar)
\end{multline*}
We proceed to prove \rref{thm:safe} for waypoint-following safety.
Per standard notation, the formula on the bottom (conclusion) is \emph{valid} if all formulas on top (premisses) are \emph{valid}:
\[
\cinferenceRule[loop|{\sf{loop}}]{}
{
\linferenceRule[formula]
{ P \limply J
& J \limply Q
& J \limply\dbox{\alpha}{J}
}
{P \limply \dbox{\prepeat{\alpha}}{Q}}
}{}
\]
The first two premisses, which prove automatically in \KeYmaeraX, say the invariant $J$ holds initially and the postcondition follows from the invariant $J$,  so staying inside $J$ is safe.

The main task is the third premiss: loop body $\alpha \equiv \ctrl;\plant$ preserves the invariant $\linv$.
The key conditions are $\planreq$ and $\admiss$; standard (\textsf{auto}matic) \dL proof steps reduce this to showing that $\linv$ holds again after running the plant for time $t \leq T$:
\[
\cinferenceRule[auto|{\sf{auto}}]{}
{
\linferenceRule[formula]
{\linv,\planreq,\admiss \limply \dbox{\plant}{\linv}}
{\linv \limply \dbox{\ctrl;\plant}{\linv}}
}{}
\]
It remains to show the premiss of \textsf{auto}: $\planreq$ and $\admiss$ imply that  $\annul$, $\controllableGoalDist(\vhvar,\vvar,\Bvar,\xgvar,\ygvar)$, and $\controllableGoalDist(\vlvar,\vvar,\Avar,\xgvar,\ygvar)$  \emph{continue to hold} throughout the $\plant$, which can be proved in \dL using \emph{differential induction}~\cite{DBLP:books/sp/Platzer18}.
\KeYmaeraX helps prove invariants, but identifying them takes human ingenuity.
For example, $\controllableGoalDist(\vhvar,\vvar,\Bvar,\xgvar,\ygvar)$ is a natural choice, but because it is not inductive, we generalize it by hand in the full proof.

This completes the safety proof, showing that requirements $\planreq$ and $\admiss$ guarantee that the robot obeys speed limits.
We describe liveness next, which will reuse loop invariant $\linv$.

\subsection{Proving Liveness}
We show $\admiss$ also allows the robot to reach the goal within speed limits $[\vlvar,\vhvar]$ by choosing correct acceleration $\avar$.
The proof starts with the \textsf{loop} rule with invariant $\linv$ as in \rref{thm:safe}, after which it remains to show:
\[\linv \land \vvar{>}0 \land \xgvar{>}0 \limply \ddiamond{\prepeat{(\ctrlliv;\plant)}}{(v \in [\vlvar,\vhvar]~\land~\enorm{(\xgvar,\ygvar)} \leq \veps)}\]
Aside from the \emph{invariants}, the key insight for liveness is a \emph{progress function} which decreases as the waypoint approaches the origin~\cite{DBLP:conf/fm/SogokonJ15}.
There are multiple strategies to arrive at the goal within speed limit; for simplicity, we first enforce the speed limit $\vvar \in [\vlvar,\vhvar]$ with appropriate acceleration choices and then maintain it until reaching the waypoint.
This strategy splits the proof into the following two questions:
\begin{equation}\label{eq:dv}
\linv \land \vvar>0 \land \xgvar>0 \limply \ddiamond{\prepeat{(\ctrlliv;\plant)}}{(\vvar \in [\vlvar,\vhvar])}
\end{equation}
\begin{multline}\label{eq:reachwaypoint}
\vvar>0 \land \vvar \in [\vlvar,\vhvar] \land \annul \limply\\ \ddiamond{\prepeat{(\ctrlliv;\plant)}}{(\vvar \in [\vlvar,\vhvar] \land \enorm{(\xgvar,\ygvar)} \leq \veps)}
\end{multline}

To prove \rref{eq:dv}, we pick the acceleration for $\ctrlliv$ in each of three situations:
\begin{inparaenum}[\it i)]
\item\label{case:tooslow} if the robot is too slow ($\vvar<\vlvar$), it should speed up (pick $\avar=\Avar$),
\item\label{case:inlim} if the speed is in the limits ($\vvar \in [\vlvar,\vhvar]$) it maintains speed (pick $\avar=0$), or
\item\label{case:toofast} if it is too fast ($\vvar>\vhvar$), it slows down (pick $\avar=-\Bvar$).
\end{inparaenum}
\newcommand{\branchinv}[1]{J_{#1}}
In all three cases, the invariant $\vvar>0 \land \annul$ is proved to be preserved throughout $\plant$, so we soundly assume it in the proof of~\rref{eq:reachwaypoint}.
The progress functions are case specific: e.g., in \rref{case:tooslow}, the progress function $g \equiv \vlvar-\vvar$, for the gap to the speed limit, decreases as the robot speeds up.
Conversely, the progress function for \rref{case:toofast} is $g \equiv \vvar-\vhvar$.

\begin{figure}[h!]
\centering
\includegraphics[width=0.3\textwidth]{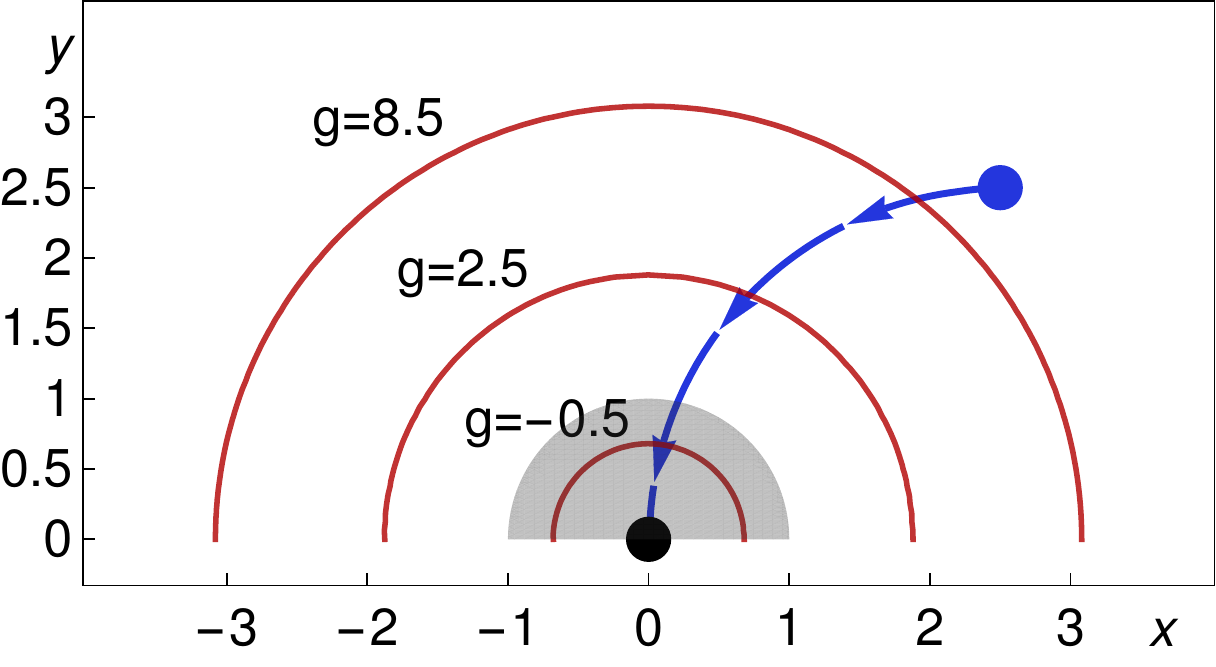}
\caption{Level sets of the progress function $g$ (in red).}
\label{fig:stagingwaypoint}
\label{fig:ode4}
\end{figure}

Once the speed limit is achieved (\rref{case:inlim}), the robot progresses toward the waypoint (\rref{eq:reachwaypoint}) at a constant velocity ($\avar = 0$).
The proof uses the progress function $g=\xgvar^2+\ygvar^2-\veps^2$, i.e., the (squared) Euclidean distance to the goal region.
The intuition for this progress function is shown in~\rref{fig:stagingwaypoint}. The value of $g$ is positive outside the goal region, strictly decreases along the trajectory, and is negative in the goal region.

These are the crucial ingredients in the proof of liveness, which we have formally proved in \KeYmaeraX. 

\subsection{Proof Effort and Automation}
\label{sec:effort}
User interactions are usually required for significant \dL theorems like \rref{thm:safe} (279 interactions) and \rref{thm:liveness} (589 interactions).
User insight was mainly needed to choose invariants and progress functions for loops and ODEs.
Most interactions are simplifications to help the automation.
Automation handled most steps: 53,883 for safety, 225,607 for liveness.
The final \KeYmaeraX proof scripts run automatically in 23s and 73s on a 2.4GHz i7 with 16GB memory.

\section{Implementation: Simulation}
\label{sec:validation}
In this section, we fulfill the goal of extending verification to the level of simulation in AirSim.
We use the \VeriPhy tool to synthesize an automatically-verified monitor containing both \emph{controller} and \emph{plant} monitor conditions.
The controller monitor condition corresponds to $\planreq$ and $\admiss$ in \rref{sec:robot-model}: any control decision satisfying these conditions is allowed and is guaranteed safe by \rref{thm:safe}, else the verified fallback is invoked.
\VeriPhy guarantees that the monitor is safe down to its machine code implementation, regardless what decisions are made by the external controller, \emph{so long as} the plant monitor conditions are satisfied, which is the case so long as the differential invariant of \rref{sec:robostage} (the premiss of \textsf{auto}) holds for the sensed values.
When the plant monitor conditions fail, safe braking is employed, albeit without the strong guarantee available in the other cases except with extra assumptions \cite{DBLP:journals/fmsd/MitschP16}.

\begin{figure*}[tb]
\centering
\begin{minipage}[b]{2in}\centering
\begin{tikzpicture}[->,draw,thick,yscale=0.9,inner sep=1pt]
\tikzstyle{arc}=[circle,draw]
\tikzstyle{line}=[rectangle,draw]
\tikzstyle{way}=[diamond,draw]
\node[way] (a) at (1,3) {A};
\node[way] (b) at (4,3) {B};
\node[way] (c) at (4.75,1.5) {C};
\node[way] (d) at (3.25,1.5) {D};
\node[way] (e) at (4,0) {E};
\node[way] (f) at (1,0) {F};
\node[way] (g) at (1.75,0.90) {G};
\node[way] (h) at (0.25,0.90) {H};
\node[way] (i) at (1.75,2.15) {I};
\node[way] (j) at (0.25,2.15) {J};
\draw (a) -> (b);
\draw (b) edge [bend left] node {} (c);
\draw (b) edge [bend right] node {} (d);
\draw (c) edge [bend left]node {} (e);
\draw (d) edge [bend right] node {} (e);
\draw (e) -> (f);
\draw (f) edge [bend left]  node {} (h);
\draw (f) edge [bend right] node {} (g);
\draw (h) -> (j);
\draw (g) -> (i);
\draw (j) edge [bend left]  node {} (a);
\draw (i) edge [bend right] node {} (a);
\end{tikzpicture}
\subcaption{Example mission}\label{fig:patrol-mission-plan}\end{minipage}
\begin{minipage}[b]{1.51in}\centering
\includegraphics[width=1.51in,clip,trim=0 0 0 50]{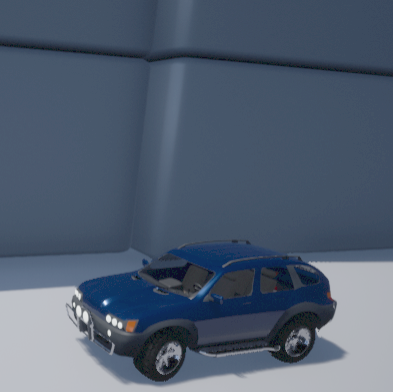}
\subcaption{Simulator}\label{fig:simulator}\end{minipage}
\begin{minipage}[b]{0.15\textwidth}\centering
\includegraphics[width=1in]{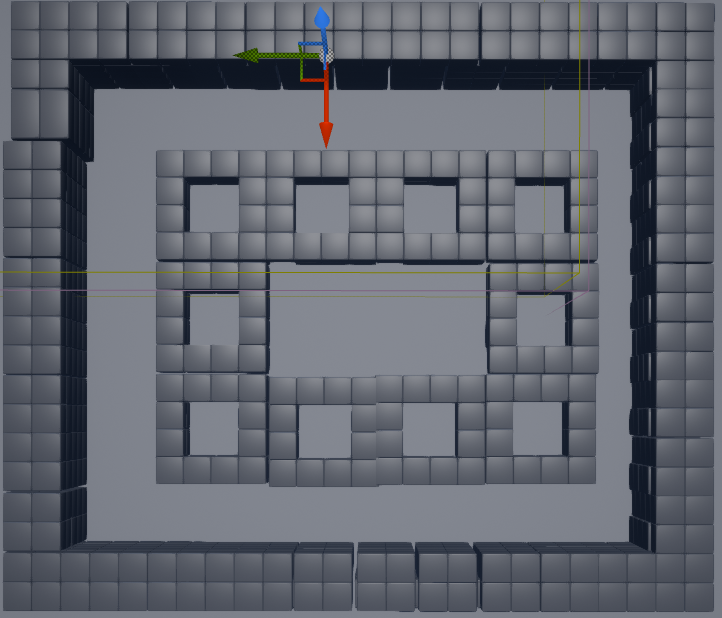}
\subcaption{Rectangle}\label{fig:rect}\end{minipage}
\begin{minipage}[b]{0.15\textwidth}\centering
\includegraphics[width=1in]{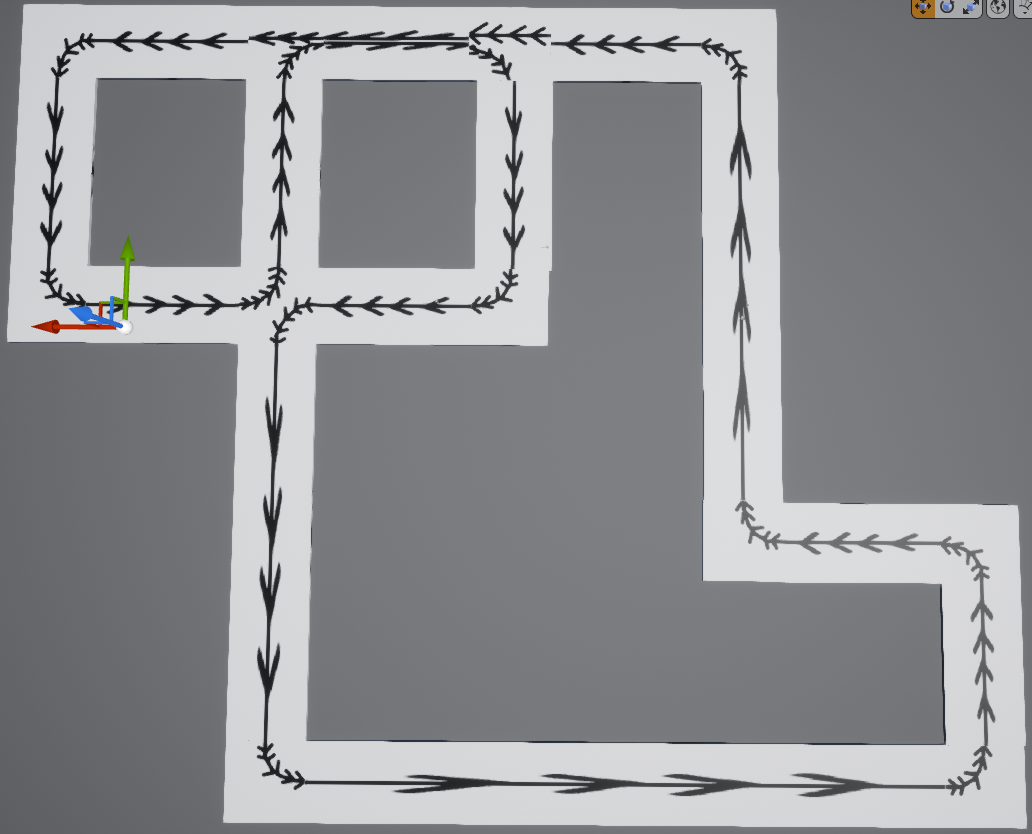}
\subcaption{Tight turns}\label{fig:turns}\end{minipage}
\begin{minipage}[b]{0.15\textwidth}\centering
\includegraphics[width=0.8in]{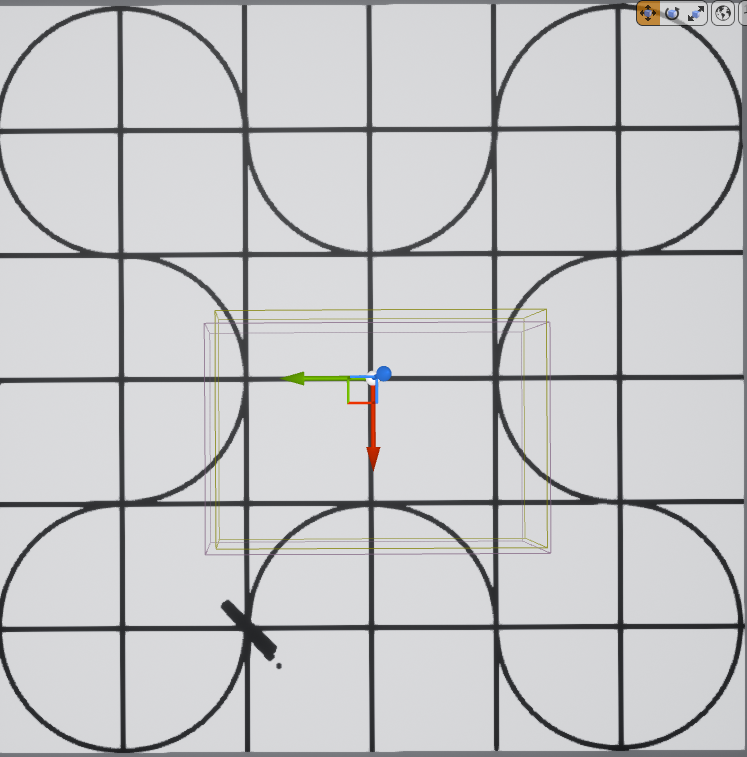}
\subcaption{Large clover}\label{fig:clover}\end{minipage}
\caption{Implementation and environments built in AirSim}
\label{fig:patrol-plan}
\end{figure*}
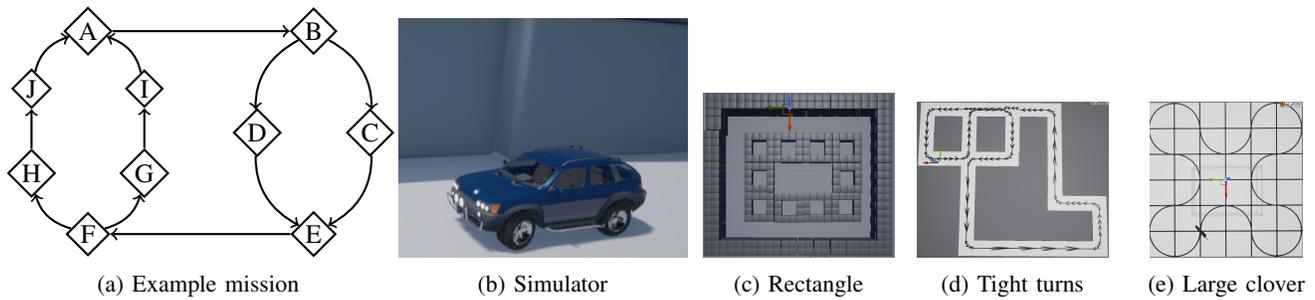

\paragraph{High-Level Plans}
Our plan data structure is a graph (\rref{fig:patrol-mission-plan}) where waypoints are connected by lines and arcs, as in the \dL model.
The evaluation uses  fixed plans (up to ${\approx}$80 segments), but our data structure also supports, e.g., nondeterministic plans (\rref{fig:patrol-mission-plan}, nodes B and F) and cyclic graphs for repeating missions,
for the sake of flexibility.
\paragraph{Feedback Control}
The high-level plan gives an ideal path to follow; the job of the low-level controller is to follow it within some tolerance.
We give two classical feedback controllers: the \emph{bang-bang} controller switches between hard-left and hard-right steering,
while the \emph{PD} scales to the discrepancy between current position and orientation vs. their target values.
We compare the low-level controllers in AirSim, using a human operating AirSim as a baseline.

\paragraph{AirSim Simulation}
We implemented our own plan representation and controllers in AirSim, $\approx$1100 lines of C++.
We built the test environments (Fig.~\ref{fig:rect},\ref{fig:turns},\ref{fig:clover}) in Unreal Editor.
\rref{fig:simulator} shows the AirSim car driving autonomously.

\paragraph{Sensing and Actuation}
AirSim does not explicitly simulate sensing and actuation error, but some implementation details of the kinematics are unknown, so actuation error must be accounted for in practice.
The tolerance $\veps$ does not include sensing error, but does account for deviation of the AirSim kinematics from ideal Dubins.
Thanks to the proofs and sensors, errors do not accumulate: if actuation is imperfect, the deviation is detected by the sensors and feedback control counters the deviation as usual.
If our monitor conditions are applied in systems where sensors accumulate drift over time, it would not obviate the need to account for those drifts.

\paragraph{Results}
We assess monitor compliance and safety of each controller.
We assess liveness indirectly by checking how \emph{quickly} the goal is reached.
We assess compliance and safety directly: a successful simulation should comply with the monitor conditions (especially the safety-critical plant monitor conditions) a large majority of the time  and have no safety violations.
Our three AirSim environments are shown in \rref{fig:patrol-plan}.
These environments cover medium turns at medium speed (\rref{fig:rect}), tight turns at low speed (\rref{fig:turns}), and wide curves at high speed (\rref{fig:clover}).
We simulated bang-bang and PD controllers of different speeds driving each environment (\rref{tab:results}), with amateur human pilots as a baseline.

The car completes every environment, except ``Clover'' where bang-bang control fails to track long curves.
As promised, there were no safety violations.
The controller monitor condition has few failures.
The plant monitor condition fails more often, but rarely enough that the car completes the drive.
The plant monitor failed more since our bloated ideal dynamics are simpler than the AirSim physics.
In general, the failure rate increases the more physics differs from Dubins.

\begin{table*}
\caption{Average speed, Monitor failure rates, for AirSim and human driver in each environment}
\label{tab:results}
\centering
\begin{tabular}{l|l|l|l|l|l|l|l|l|l|l|l|l|l|l|l}
\bottomrule        & \multicolumn{5}{|c|}{Avg.\ Speed (m/s)} & \multicolumn{5}{c|}{Ctrl Fail.} & \multicolumn{5}{c}{Plant Fail.} \\
World   & BB   & \textbf{PD1}   & PD2   & PD3     & Human     & BB     & \textbf{PD1}      & PD2    & PD3    & Human  & BB     & \textbf{PD1}    & PD2   & PD3 & Human \\\toprule
Rect    & 4.3  & \textbf{6.32}   & 7.16 &  12.6   & 9.92      & 0.5\%  & \textbf{0.1\%}    & 0.1\%  & 0.19\% & 1.14\% & 36.8\% & \textbf{8.23\%} & 8.5\%  & 14\%  & 41.3\%  \\
Turns   & 3.78 & \textbf{3.95}   & 4.43 &  4.69   & 9.66      & 1.0\%  & \textbf{1.0\%}    & 1.1\%  & 4.7\%  & 3.61\% & 18.6\% & \textbf{3.95\%} & 6.8\%  & 11\%  & 21.1\%  \\
Clover  & X    & \textbf{29.5}   & 29.5 &  29.5   & 28.9      & X      & \textbf{0.2\%}    & 0.2\%  & 0.19\% & 0.29\% & X      & \textbf{66\%}   & 66\%   & 66\%  & 48\%
\end{tabular}
\vspace{-1ex}
\end{table*}
We ran the tests with $\veps = 1\textrm{m},$ which was small enough to stress-test the controllers.
For our purposes, the exact value of $\veps$ is less important than the fact that safety is guaranteed for all values of $\veps$.
The bang-bang controller exhibits tracking error at speed and so did not complete the Clover track.
The slower PD controller (PD1, in bold) had the best (lowest) overall error rate.
The human and the remaining controllers had high plant failure rate on the ``Clover'' level due to tracking error.
PD control (particularly PD3) had speed and monitor failure rates competitive with the human baseline.
The bang-bang controller's rough steering increased its plant failure rate.

While complete model compliance is a challenge, well-tuned controllers came close in all environments, \emph{even though} the model is simple.
The crucial insight is that the bloated model allows realistic imperfections in actuation, and that the \emph{untrusted controller} makes steering and acceleration choices that satisfy its monitor condition.
Most of the time, formal guarantees apply because both monitor conditions are satisfied.
The plant's monitor condition detects the few cases where guarantees do not apply, engages the fallback action, and then returns to normal control without any actual safety violation.
The monitor furthermore helps us or any other developer identify and reduce the remaining non-compliant cases.

\section{Conclusions and Lessons}
We cast a formally verified safety net that provides end-to-end verification guarantees for 2D Dubins waypoint-following.
We developed a \dL model, proved it safe and live in \KeYmaeraX, then synthesized a verified monitor with ModelPlex, and synthesized verified machine code with \VeriPhy.
The resulting safety net ensures safety even with unverified robot software so long as \emph{plant} assumptions hold and collision-free plans are provided.
We simulate the system in AirSim with several controllers; our aim was not to innovate in controller design, but to show that monitors generated from \dL proofs can be applied in realistic scenarios, thanks largely to the use of verified tolerances in the model and proof.
The evaluation showed that our simple tolerance-based model did not hinder applicability, because even realistic simulations look like Dubins at a distance.
Our simple model greatly eased formal verification.
Just as we improved on prior models~\cite{DBLP:conf/pldi/BohrerTMMP18}, future work can verify more sophisticated models to improve compliance or reduce the tolerance $\veps$.

The second major direction for future work is to apply \VeriPhy on real robot hardware, and as a development aid for novel robots.
Because the \KeYmaeraX proofs are significantly more complex than the sketches presented here (see~\rref{sec:effort}), \VeriPhy's reusability is essential to make it practical as a development tool. We are presently in the process of reusing our current monitors as-is on a hardware platform that follows Dubins paths without changing the proofs, and subsequently intend to pursue more challenging motion scenarios that violate our model's assumption.

\bibliographystyle{IEEEtran}
\bibliography{paper}

\bibliographystyle{plainnat}
\bibliography{root.bib}

\end{document}